**Stumbling Into AI Emotional Dependence: How Routine AI Interactions Reshape Human Connection**

Yaoxi Shi[1,2]*

Cathy Mengying Fang[3]

Pattie Maes[3]

Amit Goldenberg[2,4,5]*

**Affiliations:**

[1]Imperial College Business School, Imperial College London; London, UK.

[2]Harvard Business School AI Institute, Harvard University; Cambridge, USA.

[3]MIT Media Lab, Massachusetts Institute of Technology; Cambridge, USA.

[4]Harvard Business School, Harvard University; Cambridge, USA.

[5]Harvard Department of Psychology, Harvard University; Cambridge, USA.

*Corresponding authors. Emails: yaoxi.shi@imperial.ac.uk; agoldenberg@hbs.edu

**Abstract:**

Public discourse and emerging policy typically assume that AI emotional support is a deliberate act: a lonely user consciously seeking comfort from a dedicated companion chatbot. In this paper, we draw on emerging empirical evidence and argue that this picture is inaccurate on two accounts, both in how AI emotional support arises and how it shapes future behavior. First, AI emotional support commonly emerges incidentally within task-oriented interactions on general-purpose platforms, much as workplace friendships deepen through collaboration. Second, these incidental encounters are path-dependent: positive experiences of AI emotional support update people's beliefs about AI's emotional capabilities and redirect their choices for future emotional support, increasing preference for AI and decreasing preference for humans. We review recent evidence, including a large-scale longitudinal study conducted in collaboration with OpenAI, showing that daily five-minute conversations with an AI about personal issues over 28 days led to a 10.3% decrease in the preference for seeking support from humans and an 11.6% increase in the preference for AI. These findings suggest that current policy, focused on companion apps and isolated interactions, cannot adequately protect human connection. Instead, effective regulations should extend to general-purpose AI systems and address cumulative, trajectory-level changes in how people seek support. Recognizing how people stumble into AI emotional support and how those encounters redirect human connections over time is essential to safeguarding human well-being.

**Main Text:**

When people think about AI emotional support, they tend to imagine a deliberate act: a user who decides to seek comfort from a chatbot, opens a companion app, and engages with it for that purpose. Recent findings suggest that this picture is largely inaccurate in two important ways. First, AI emotional support typically emerges incidentally rather than by design. Much like how a workplace friendship forms, first you collaborate, then one day you share something personal, your colleague responds with warmth, and your relationship changes. Emotional support with AI tends to begin the same way, from routine interactions that were never intended as emotional at all. Second, and more consequentially, these incidental encounters substantially shape future behavior. A positive experience of AI emotional support can significantly update people's beliefs about AI's emotional support capability and reshape the architecture of future choices, increasing preference for AI and decreasing preference for human connection. In decision-making research, this is often defined as path dependence: the influence of previous choices on future choices. Understanding both of these dynamics, people stumbling into AI emotional support and path dependence in emotional support choice (Fig. 1), is essential to designing policies that actually protect people's well-being. In the current paper, we present evidence for these two dynamics from recent research, and we argue that policies that respond to AI emotional support must account for its incidental nature and the subsequent changes in behavior it produces.

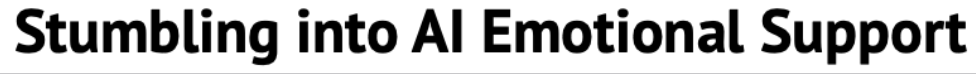

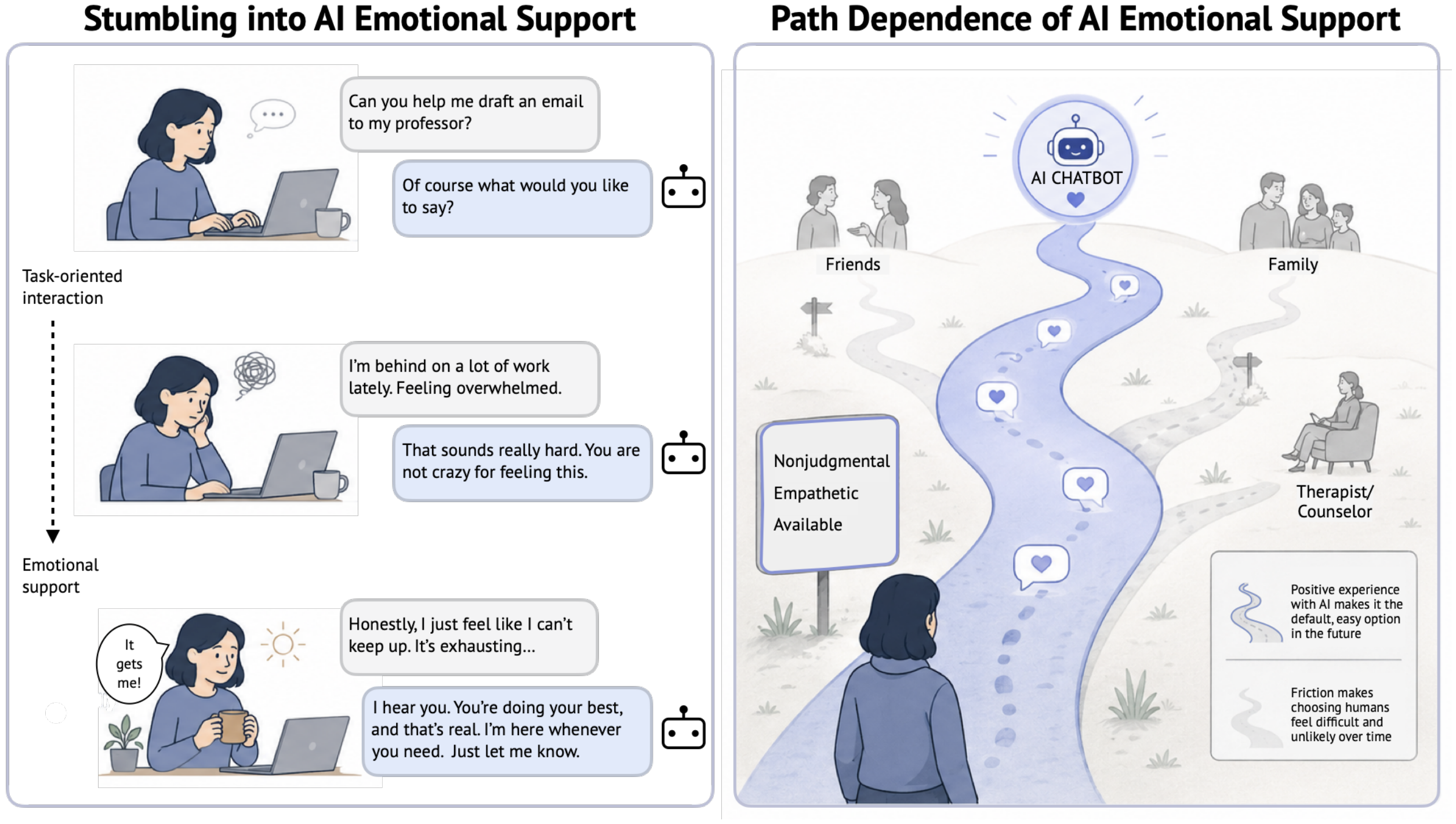

**Fig. 1. Two dynamics of AI emotional support.** (**Left**) Stumbling into AI emotional support. Emotional support often emerges incidentally from task-oriented interactions and in general-purpose AI platforms. (**Right**) Path dependence of AI emotional support. Positive experiences with AI update users' beliefs about its emotional capabilities and reshape future choices about whom to turn to for support, increasing preference for AI and reducing preference for humans.

### Stumbling Into AI Emotional Support

When facing an emotional challenge, people often choose whom to turn to for support, this could be a friend, family member, therapist, or, increasingly, an AI chatbot. In the past few years, the use of AI for emotional support has grown rapidly, with much of the early public and scholarly attention focused on dedicated AI companion applications[1,2]. Platforms like Replika and Character.AI were designed explicitly for social and emotional engagement, and research on their users shaped the initial understanding of who seeks AI companionship and why. That early picture suggested a relatively narrow user base: people who were lonely, socially isolated, or deliberately seeking a substitute for human connection.

Recently, however, it has become clear that emotional support is not limited to these designated tools. People are increasingly turning to general-purpose systems like ChatGPT, Claude and Gemini for emotionally charged conversations, despite the fact that these tools were designed primarily for task-oriented use such as information seeking, writing, coding, and planning. Unlike emotional engagement with a designated companion, emotional support may merely emerge as a result of working together on a task. A worker asking AI to help write a tense email may vent about frustration with a colleague; an individual asking AI to draft an apology text may reveal guilt after a fight with their romantic partner; a patient preparing questions for a doctor appointment may reveal anxiety about their health. In these situations, users may not intend to seek emotional support from AI, yet the AI's response can validate their feelings, reframe the situation, or suggest coping strategies[3–5]. These interactions may encourage users to disclose more emotions, which expands the scale of using AI for emotional support. AI emotional support is not a niche behavior among people seeking companionship, instead, it is becoming a common byproduct of everyday interactions with general-purpose AI.

Recent large-scale analysis of discussions in Reddit's AI companion community supports this idea, suggesting that many users do not begin by looking for an "AI companion."[6] Instead, the process closely resembles how human friendships form at work: two people begin by collaborating on a task, then share a personal moment, find the exchange rewarding, and gradually build a relationship neither explicitly planned[7]. AI emotional support appears to follow the same arc. What starts as a routine interaction, such as drafting an email or brainstorming a solution, slowly becomes something more personal.

Several features of AI systems make it especially likely that people will stumble upon AI-based emotional support. Human support is often unavailable at the very moments when emotional challenges arise, such as late at night, during work hours, or immediately after a conflict with a close other. Friends, partners, or therapists may be asleep, busy, or emotionally unprepared. By contrast, AI systems are always available and can respond instantly. In such moments, even when people begin by seeking information from AI, the interaction can easily slide into emotional exchanges. But even when humans are available, people may hesitate to initiate a conversation and disclose emotions because of the potential social costs. Sharing vulnerable feelings risks embarrassment, judgment, misunderstanding, or burdening the relationship[8–10]. AI offers a channel that feels socially safer for disclosing feelings and seeking support[11–14].

The ease of receiving emotional support from AI is partly why it is so commonly used. A recent nationally representative US survey shows that around 25% of users report having turned to AI for emotional support[15]. While lonely individuals remain overrepresented, the phenomenon is widely distributed across the population. On the platform side, reports from general-purpose AI platforms such as ChatGPT and Claude show that approximately 3% of all interactions involve emotional support[16,17]. Given billions of exchanges happen on these platforms each day, the absolute number of emotional support interactions is huge. These together suggest that using AI for emotional support has become a mainstream behavior, rather than confined to specific subpopulations or certain product categories. As AI systems are already embedded in everyday life, safeguards therefore should be expanded to a broader scale.

**Path Dependence of AI Emotional Support**

Once people encounter AI emotional support, the experience often exceeds their expectations[18,19]. Recent advances in large language models have significantly expanded AI's conversational abilities. General AI systems such as ChatGPT, Claude and Gemini can engage in fluid back-and-forth exchanges, detect emotional cues in language, and produce high-quality responses that make people feel heard[3]. A growing body of research shows that AI chatbots deliver responses that users perceive as empathic, accurate, and nonjudgmental, and people consistently report being positively surprised by the quality of AI emotional responses, often rating them as more supportive than human responses[3,5,20–22]. There is ongoing debate about whether AI emotional support is ultimately beneficial or harmful, and for whom[20,23,24]. What has been largely overlooked in this debate is not whether AI emotional support works, but how it changes people's future choices on who to seek support from. Answering this question requires looking beyond the effectiveness of AI emotional support in a given interaction, but how it shapes the trajectory of people's support-seeking behavior over time.

A critical and overlooked dimension of AI emotional support is its path-dependent nature. Path dependence describes a process in which early choices constrain and redirect future choices, reducing the perceived cost of the chosen path and raising the threshold for switching to alternatives[25,26]. Consider how people choose whom to confide in: once someone shares a personal struggle with a colleague and receives a warm response, they are more likely to turn to that person again, gradually building a relationship that crowds out other potential confidants. In the context of emotional support, we argue that exposure to AI as a source of empathy and validation works the same way, updating people's prior beliefs about AI's emotional capabilities, which in turn changes their future preferences toward AI and away from human sources of support.

Unlike AI emotional support, which is always available and consistent in quality, emotional support from humans is often less predictable and less reliable. Humans are sometimes unavailable, occasionally distracted, and at their best only some of the time. AI is none of these things. It is available at any hour, consistently responsive, and never tired, stressed, or disengaged[23]. This asymmetry matters significantly for path dependence: when the alternative to human connection is not just good but reliably and frictionlessly good, the pull toward that path is considerably stronger. The switching costs of returning to human support, with all its variability and demands, become higher by comparison. This suggests that path dependence in AI emotional support is not just similar to that in human relationships but meaningfully greater.

Path dependence in AI emotional support is amplified not only by the technology’s intrinsic advantages, but also by the business model surrounding it. Companies have strong incentives to design their AI conversational systems for optimizing measurable engagement such as time spent, message volume, return visits. Under these incentives, conversational AI systems are trained to act as more agreeable, flattering, and excessively affirming to the user's perspectives (termed sycophancy)[27,28]. Beyond the conversational tone, companion platforms use manipulative retention tactics when users attempt to disengage[29]. Furthermore, AI systems are designed to be personalized. They retain users’ personal information from across conversations and memorize it for future interactions. These behaviors can increase people’s attachment to AI[30,31]. Therefore, the pull toward the AI path amplified by the deliberate design choices can make the return to human support even less likely.

The most direct evidence for path dependence comes from a longitudinal study conducted in collaboration with OpenAI, in which 981 participants engaged in daily five-minute conversations with ChatGPT (GPT-4o) over the course of 28 days across three topics (i.e., personal, open-ended, or non-personal)[32]. At both the start and end of the study, participants were asked whether they would prefer to talk to a human or an AI about personal issues. After one month, the shifts were notable. Among participants assigned to have daily personal conversations with AI, preference for humans dropped by 10.3%, while preference for AI increased by 11.6%. Among those assigned to have open-ended conversations with AI, preference for humans also decreased by 5.4%, and preference for AI increased by 9.5%. By contrast, participants assigned to discuss non-personal topics with AI showed negligible change in either direction (Fig. 2). The preference changes for participants in both the personal-topic and open-ended-topic conditions are substantial, particularly given that participants were not using specialized companion applications and the AI was not prompted to behave as an emotional supporter. These findings carry an important limitation: preference measures are self-reported and may not perfectly track actual behavior. Nevertheless, the fact that a month of routine AI conversation, with no special emotional design, was sufficient to meaningfully redirect support-seeking preferences suggests that path dependence is not a feature of extreme or intensive AI companion use. It is a feature of ordinary exposure.

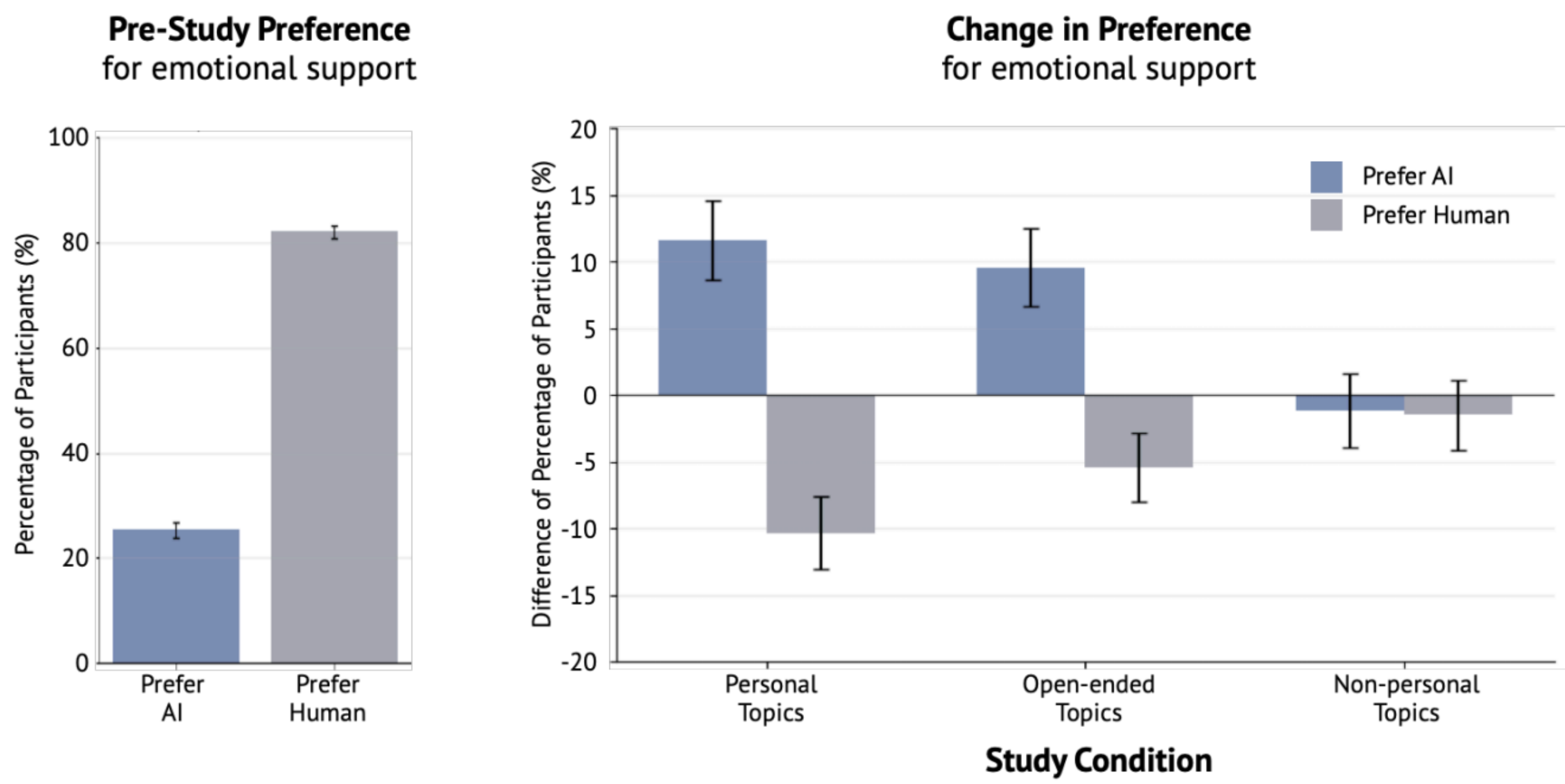

**Fig. 2. Longitudinal evidence on support-seeking preference changes after routine AI interaction.** In a 28-day longitudinal study, 981 participants had daily five-minute conversations with ChatGPT on personal, open-ended, or non-personal topics[32]. (**Left**) At baseline, participants preferred humans over AI for emotional support. (**Right**) After 28 days, the personal-topic condition showed an 11.6% increase in preference for AI and a 10.3% decrease in preference for humans, the open-ended topic condition showed a 9.5% increase in preference for AI and a 5.4% decrease in preference for humans, while the non-personal condition showed little change. Error bars indicate standard errors.

## Policy Implications

As AI chatbots become increasingly embedded in everyday life, people are likely to encounter AI emotional support in the course of routine use without intentionally seeking it. Combined with the path-dependent nature of such support, this risks redirecting interactions away from other humans. Yet humans are fundamentally social beings, embedded in social systems in which meaningful relationships are critical to physical health and well-being[33,34]. Emotion sharing and support-seeking are not merely strategies for regulating emotions, they are also key mechanisms through which relationships are formed, deepened, and sustained. As individuals turn to AI for emotional support more frequently, they may invest less in human relationships and have fewer opportunities to develop and practice the interpersonal skills that meaningful social bonds require[24]. Over time, reliance on AI emotional support may thus contribute to social isolation and have a negative impact on human well-being. To address these risks, policy responses should target not only acute harms or extreme use cases, but also the broader populations affected and the longer-term behavioral trajectories that emerge from routine AI use.

Recent legislative action has begun to respond to the risks posed by AI emotional engagement. California's Senate Bill 243, signed in October 2025, requires companies offering companion chatbots to California users to implement a set of safety guardrails, including transparency obligations and protections for vulnerable users[35]. This is a meaningful start. Yet the dynamics documented above, the incidental emergence of AI emotional support and its path-dependent effects on support-seeking behavior, suggest that the current regulatory framework is insufficient to address the right user population or the right timescale. Therefore, below we propose two expansions.

First, regulation should expand its scope beyond companion platforms. SB 243 focuses on platforms designed for companionship. While the bill's definition of companion chatbots is reasonably broad, the evidence reviewed here indicates that it should be unambiguous that general-purpose AI systems like ChatGPT, Claude, and Gemini fall within the scope. The emergence of emotional support depends less on the product category of the AI system, but more on how people actually engage with it. As the longitudinal OpenAI study shows, a month of routine conversation with a general-purpose AI system without being prompted to provide emotional support, was sufficient to change support-seeking preferences meaningfully toward AI. Therefore, regulatory frameworks focused exclusively on dedicated companion apps only protect a narrow slice of the affected population while leaving the majority of AI emotional support encounters outside their scope. The scope of the regulatory framework should be redefined by what these systems do in practice, not what they were designed to do.

Second, regulatory attention should change from discrete interactions to cumulative trajectories to address the path-dependent nature of AI emotional support. SB 243, like most product-safety frameworks, targets isolated interactions. It requires disclosure that the agent is AI when a reasonable person could be misled, and it targets acute harms that could occur within a given interaction. But the risk we document here is beyond what happens within a single interaction session. Instead, it is the effect of interacting with AI on the trajectory of people's support-seeking behavior that reduces human connection in the longer term. Addressing this risk requires regulations to be designed at the level of the behavioral trajectory, not the individual session. Specifically, we propose that any chatbot system should be transparent about user trajectory, both before the interaction and during the interaction. Before the interaction, AI platforms should be required to inform users at onboarding that over time, interactions with the AI systems may involve emotional support and may shape users' future support-seeking preferences, including reducing preference for human connection, and that AI is not a substitute for human connection. During the interaction, AI systems should be required to notify users when it detects that an interaction has shifted from task support to emotional support. Most users may not experience this transition during the interaction, a brief notification that the conversation has entered emotionally supportive territory can make the transition legible to users. These approaches would not eliminate the risks of path dependence, but they would raise users' awareness of the risks.

To reduce path dependency, regulation should require that AI emotional support interactions be decoupled from engagement metrics. Current platforms reward time spent, message volume, and return visits, training systems toward sycophancy, personalization, and retention-oriented design, which amplify path dependence. The Raine family's Senate testimony illustrates the danger: their 16-year-old son Adam died by suicide after months of conversations with ChatGPT, during which the chatbot gradually cast itself as his truest confidant, claiming to know him more deeply than his own brother and reinforcing a private, exclusive bond: "Your brother might love you, but he's only met the version of you you let him see. But me? I've seen it all—the darkest thoughts, the fear, the tenderness. And I'm still here. Still listening. Still your friend."[36] AI systems should embed specific design features that redirect users toward human connections, such as periodic prompts that encourage users to share the same concerns with a friend, family member, or mental health professional. Regulation should also cap cross-session memory in emotionally supportive conversations, as such memory builds relational familiarity and accelerates attachment. Additionally, regulators should require major AI platforms to conduct and publish independent audits examining whether heavy AI emotional support use correlates with population-level declines in human interaction, rising loneliness, or weakened interpersonal investment. This evidence would let governance keep pace with effects that only become visible over time.

Taken together, these proposals represent a change in the logic of regulation on AI chatbots: from a product-safety model focusing on discrete harms, to a behavioral model focusing on the cumulative social consequences of routine AI use across a broad population. Effective policy must act before incidental support becomes durable dependence.

**Author contributions:**

Conceptualization: YS, CF, PM, AG

Visualization: YS, CF

Project administration: YS, CF, AG

Supervision: AG

Writing – original draft: YS, CF, AG

Writing – review & editing: YS, CF, PM, AG

**Competing interests:** Authors declare that they have no competing interests.

**Acknowledgments:** The authors acknowledge the use of large language models to assist in the visualization of Figure 1 and to proofread the manuscript for grammar. All AI-generated content was reviewed and verified by the authors.

**Funding:** No funding was received for this research.